\documentclass[12pt]{article}

\usepackage{bbm,epic,eepic,framed,epsf,latexsym,amsmath,amssymb,amscd,mathrsfs,slashbox }
\usepackage[all]{xy}

\usepackage{graphicx}
\usepackage[pdftex,colorlinks]{hyperref}

\usepackage[margin=1.2in]{geometry}
\newtheorem{theorem}{Theorem}[section]
\newtheorem{lemma}[theorem]{Lemma}

\newtheorem{note}[theorem]{Note}

\newtheorem{counter-example}[theorem]{Counter example}
\newtheorem{proposition}[theorem]{Proposition}
\newtheorem{open question}[theorem]{Open question}
\newtheorem{corollary}[theorem]{Corollary}

\newtheorem{claim}{Claim}

\newcommand{\ignore}[1]{}

\title{Clustering is difficult only when it does not matter\thanks{Credit for this title goes to Tali Tishby who stated this in a conversation with one of us many years ago.}}

\begin{document}

\author{Amit Daniely \thanks{Department of Mathematics, Hebrew
                 University, Jerusalem 91904, Israel. Supported in part by a binational Israel-USA grant 2008368. amit.daniely@math.huji.ac.il}
                 \and Nati Linial \thanks{School of Computer Science and Engineering, Hebrew University, Jerusalem 91904, Israel. Supported in part by a binational Israel-USA grant 2008368.  nati@cs.huji.ac.il}
                 \and Michael Saks \thanks{Department of Mathematics, Rutgers University, Piscataway, NJ 08854.  Supported in part by NSF under
grant  CCF-0832787 and by a binational Israel-USA grant 2008368. saks@math.rutgers.edu.}}

\maketitle

\setcounter{page}{0}

\thispagestyle{empty}
\maketitle

\begin{abstract}
Numerous papers ask how difficult it is to cluster data. We suggest that the more relevant and interesting question is how difficult it is to cluster data sets {\em that can be clustered well}. More generally, despite the ubiquity and the  great importance of clustering, we still do not have a satisfactory mathematical theory of clustering. In order to properly understand clustering, it is clearly necessary to develop a solid theoretical basis for the area. For example, from the perspective of computational complexity theory the clustering problem seems very hard. Numerous papers introduce various criteria and numerical measures to quantify the quality of a given clustering. The resulting conclusions are pessimistic, since it is computationally difficult to find an optimal clustering of a given data set, if we go by any of these popular criteria. In contrast, the practitioners' perspective is much more optimistic. Our explanation for this disparity of opinions is that complexity theory concentrates on the worst case, whereas in reality we only care for data sets that can be clustered well.

We introduce a theoretical framework of clustering in metric spaces that revolves around a notion of "good clustering". We show that if a good clustering exists, then in many cases it can be efficiently found. Our conclusion is that contrary to popular belief, clustering should not be considered a hard task.

\end{abstract}

\newpage

\section{Introduction}
{\em Clustering} is the task of partitioning a set of objects in a meaningful way. Notwithstanding several recent attempts to develop a theory of clustering (e.g. \cite{AckBen-David08, BalcanBlVem08, Kleinberg02}), our foundational understanding of the matter is still quite unsatisfactory.

The clustering problem deals with a set of objects $X$ that is equipped with some additional structure, such as a dissimilarity (or similarity) function $w:X\times X\to [0,\infty)$. Informally, we are seeking a partition of $X$ into clusters, such that objects are placed in the same cluster iff they are sufficiently similar. Here are some concrete popular manifestations of this general problem.
\begin{enumerate}
\item A very popular optimization criterion is $k$-means. Aside from $X$ and $w$ one is given an integer $k$. The goal is partition $X$ into $k$ parts $C_1,\ldots,C_k$ and find a {\em center} $x_i \in C_i$ in each part so as to minimize $\sum_i \sum_{y \in C_i} w^2(y, x_i)$. Other popular criteria of similar nature are
$k$-medians, min-sum and others.
\item Many clustering algorithms work ``bottom up". Initially, every singleton in $X$ is considered as a separate cluster, and the algorithm proceeds by repeatedly merging nearby clusters. Other popular algorithms work ``top down": Here we start with a single cluster that consists of the whole space. Subsequently, existing clusters get split to improve some objective function.
\item Several successful methods use spectral methods. One associates a matrix (e.g. a Laplacian) to $(X,w)$, and partitions $X$ according to the eigenvectors of this matrix.
\end{enumerate} 
Approaches to the clustering problem that focus on some objective function, usually result in $NP$-hard optimization problems. Consequently, most existing theoretical studies concentrate on designing approximation algorithms for such optimization problems and proving appropriate hardness results. 

However, the practical purpose of clustering is {\em not} to optimize such objectives. Rather, our goal is to find a meaningful partition of the data (provided, of course, that such a partition exists). The point that we advocate is that a satisfactory theory of clustering, should start with a definition of a good clustering and proceed to determine when a good clustering can be found efficiently. In this paper, we follow this approach when the underlying space in a metric\footnote{The assumption that $d$ is a metric is not too strict. E.g., much of what we do applies even if we weaken the triangle inequality to $\lambda\cdot d(x,z) \le d(x,y)+d(y,z)$ for $\lambda$ bounded away from zero.} space.

This perspective leads to conclusions which are at odds with common beliefs regarding clustering. This applies, in particular, to the computational hardness of clustering. The infeasibility of optimizing most of the popular objectives led many theoreticians, to the bleak view that clustering is hard. However, we show that in many circumstances a good clustering can be efficiently found, leading to the opposite conclusion. From the practitioner's viewpoint, {\it "clustering is either easy or pointless"} -- that is, whenever the input admits a good clustering, finding it is feasible. Our analysis provides some support to this view.

This work is one of several recent attempts to develop a mathematical theory of clustering. For more on the relevant literature, see Section \ref{sec:conclusion}.

\subsection{A Theoretical Framework for Clustering in Metric Spaces}
There are numerous notions of clusters in data sets and clustering methods to be found in the literature. Although not necessarily stated explicitly, these methods are guided by an ideal (in the Platonic sense) notion of a {\em good cluster} in a space $X$. This is a subset $C \subseteq X$ such that if $x \in C$ and $y \not \in C$, then $x$ is {\em substantially} closer to $C$ than $y$ is. To rule out trivialities we usually require $C$ to be {\em big} enough. This, in particular, eliminates the possibility of trivial singleton clusters. Even more emphasis is put on problems of {\em clustering}. Here we seek partitions of the space $X$ into {\em clusters} such that every $x \in X$ is {\em substantially} closer to the cluster containing it than to any other cluster. The problem is specified in terms of a proximity measure $\Delta(x, A)$ between elements $x \in X$ and subsets $A \subseteq X$. Numerous natural choices for $\Delta(\cdot,\cdot)$ suggest themselves. For example, if $X$ is a metric space, it is reasonable to define $\Delta(x, A)$ in terms of $x$'s distances from members of $A$.

In the present paper we consider a metric space $(X,d)$ from which data points are sampled\footnote{In certain cases it is inappropriate to assume that points of $X$ are drawn at random. It is also possible that we do not know how $X$ is sampled. In such circumstances, we consider $P$ as the uniform distribution on $X$.} according to a probability distribution $P$.
The definition we adopt here is $\Delta(x,A)=E_{y\sim P}[d(x,y)|y\in A]$. Other interesting definitions suggest themselves, e.g., $\Delta '(x,A)=\inf _{y\in A\setminus \{x\}}d(x,y)$.

{\it A technical comment: The definition of $\Delta(x,A)$ depends on the distribution $P$. To simplify notations we omit subscripts such as $P$ when they are clear from the context.}

Formally, we say that $C\subset X$ is an {\bf $(\alpha,\gamma)$-cluster} for $\alpha >0,\;\gamma >1$ if $P(C)\ge \alpha$ and for (almost-)every\footnote{{\em Almost} means, as usual, that we are allowing an exceptional set of measure zero.} $x\in C,y\notin C$,

$$\Delta(y,C)\ge \gamma\cdot \Delta(x,C).$$

Likewise, a partition $\mathcal C=\{C_1,\ldots,C_k\}$ of $X$ is an {\bf $(\alpha,\gamma)$-clustering} for some $\alpha >0,\;\gamma >1$ if 
$$\Delta(x,C_j)\ge \gamma\cdot \Delta(x,C_i)$$
for every $i\ne j$ and (almost-)every $x\in C_i$ and, in addition, $P(C_i)\ge \alpha$  for every $i$.

A few technical points are in place.
\begin{itemize}
\item 
We study $(\alpha,\gamma)$-clusterings of a space as well as partitions of a space into $(\alpha,\gamma)$-clusters. We note that although these two notions are similar, they are {\bf not} identical.
\item
Our results hold if we choose instead to define $\Delta(x,A)$
as $E[d(x,y)|y\in A\setminus\{x\}]$. This definition is perfectly reasonable, but it leads to certain minor technical complications that the current definition avoids. Moreover, the difference between the two definitions is rather insignificant, since our main interest is in cases where $P(\{x\})\ll P(A)$.
\end{itemize}
Our main focus here is on efficient algorithms for finding $(\alpha,\gamma)$-clusters and clusterings. The analysis of these algorithms rely on the structural properties of such clusters. We can now present our main results. To simplify matters without compromising the big picture, we state our theorems in the case when $X$ is a given finite metric space.
\begin{theorem}\label{taste}
For every fixed $\gamma > 1, \alpha > 0$
there is an algorithm that finds all $(\alpha, \gamma)$-clusterings of a given finite metric space $X$ and runs in time $\mbox{poly}(|X|)$.
\end{theorem}

\begin{theorem}\label{dessert}
There is a polynomial time algorithm that on input a finite metric space $X$ and $\alpha > 0$ finds all $\gamma$-clusters in $X$ with $\gamma > 3$ and a partition of $X$ into $(\alpha, \gamma)$-clusters with $\gamma > 3$, provided one exists. Moreover, the latter problem is $NP$-hard for $\gamma = 5/2$.
\end{theorem}

\subsection{An overview}
Our discussion splits according to the value of
the parameter $\alpha$. When $\alpha$ is bounded away from zero we work by exhaustive sampling (e.g. as in~\cite{ArGeSaSch12}). We first sample a small set of points $S$ from the space. Since $|S|$ is small (logarithmic in an error parameter), it is computationally feasible to consider all possible partitions $\Pi$ of $S$.
To each partition $\Pi$ of $S$ we associate a clustering that can be viewed as the corresponding ``Voronoi diagram". If the space has an $(\alpha,\gamma)$-clustering $\cal C$, let $\Pi^{\ast}$ be the partition of $S$ that is consistent with $\cal C$. We show that the ``Voronoi diagram" of $\Pi^{\ast}$ nearly coincides with $\cal C$ provided that $\gamma$ is bounded away from $1$.
Concretely, Lemma \ref{lemma:dist-bound} controls the distances between points that reside in distinct clusters in an $(\alpha,\gamma)$-clustering. Together with Hoeffding's inequality this yields Lemma \ref{lemma:sampling-lemma} and Corollary \ref{cor:clustering-can-be-approximated} which show that the ``Voronoi diagram" of an appropriate partition of a small sample is nearly an $(\alpha,\gamma)$-clustering. Lemma \ref{lemma:distinct-clusterings-are-far} speaks about the collection of all possible $(\alpha,\gamma)$-clusterings of the space. It shows that every two distinct $(\alpha,\gamma)$-clusterings must differ substantially. Consequently (Corollary \ref{only_finite_number}) there is a bound on the number of $(\alpha,\gamma)$-clusterings that any space can have. All of this is then used to derive an efficient algorithm that can find all $(\alpha,\gamma)$-clusterings of the space, proving Theorem \ref{taste}.

In section \ref{sec:many_clusters} we deal with the case of small $\alpha$. This affects the analysis, since we require that the dependency of the algorithm's runtime on $\alpha$ be $\mbox{poly}(\frac{1}{\alpha})$. We show that $(\alpha,3+\epsilon)$-clusters are very simple: Such a cluster is a ball and any two such clusters that intersect are (inclusion) comparable. These structural properties are used to derive an efficient algorithm that partitions the space into $(\alpha,3+\epsilon)$-clusters (provided that such a partition exists), proving the positive part of Theorem \ref{dessert}. To match this result, we show that finding a partition of the space into $(\alpha,2.5)$-clusters is NP-Hard{, proving Theorem \ref{dessert} in full.

Lastly, in section \ref{sec:conclusion} we discuss some connection to other work, both old and new, as well as some open questions arising from our work.

\section{Clustering into Few Clusters -- $\alpha$ is bounded away from zero}\label{sec:few-clusters}
Throughout the section, $X$ is a metric space endowed with a probability measure  $P$. To avoid confusion, other probability measures that are used throughout, are denoted by $\Pr$. We define a metric $d$ between two collections of subsets of $X$, say $\mathcal C=\{C_1,\ldots,C_k\}$ and $\mathcal C'=\{C'_1,\ldots,C'_{k}\}$. Namely, $d(\mathcal C,\mathcal C')=\min P(\cup_{i=1 }^{k}C_{i}\oplus C_{\sigma(i)}')$ where $A\oplus B$ denotes symmetric difference, and the minimum is over all permutations $\sigma \in S_k$. The definition of $d(\mathcal C,\mathcal C ')$ extends naturally to the case where $\mathcal C$ and $\mathcal C '$ have $k$ resp. $l$ sets and, say $l \le k$. The only change is that now $\sigma: [l] \rightarrow [k]$ is $1:1$.

We define $\Delta$ also on sets. If $A, B \subseteq X$, we define $\Delta(A,B)$ as the expectation of $d(x,y)$ where $x$ and $y$ are drawn from the distribution $P$ restricted to $A$ and $B$ respectively.
It is easily verified that $\Delta$ is symmetric and satisfies the triangle inequality. It is usually {\em not} a metric, since $\Delta(A, A)$ is usually positive.
\begin{proposition}
For every $A,B,C\subset X$,
$$\Delta(A,B) = \Delta(B,A)~~\mbox{and}~~~\Delta(A,B)\le \Delta(A,C)+\Delta(C,B)$$
\end{proposition}

As the following lemma shows, distances in an $(\alpha,\gamma)$-clustering are fairly regular
\begin{lemma}\label{lemma:dist-bound}
Let $C_1,\ldots,C_k$ be an $(\alpha,\gamma)$-clustering and let $i\ne j$. Then
\begin{enumerate}
\item For almost every $x\in C_i,y\in C_j$, $\frac{\gamma-1}{\gamma}\Delta(y,C_i)\le d(x,y)\le \frac{\gamma^2+1}{\gamma(\gamma-1)}\Delta(y,C_i)$ \label{internal:part-1} 
\item For almost every $x,y\in C_i$, $d(x,y)\le\frac{2}{\gamma-1}\cdot\Delta(x,C_j)$\label{internal:part-2}
\end{enumerate}
\end{lemma}
{\bf Proof.} Let $x\in C_i,y\in C_j$. For the left inequality in part \ref{internal:part-1}, note that 
\begin{eqnarray*}
d(x,y) &\ge& \Delta(y,C_i)-\Delta(x,C_i)\\
&\ge& \Delta(y,C_i)-\frac{1}{\gamma}\cdot\Delta(x,C_j)\\
&\ge& \Delta(y,C_i)-\frac{1}{\gamma}\cdot[d(x,y)+\Delta(y,C_j)]\\
&\ge& \Delta(y,C_i)-\frac{1}{\gamma}\cdot[d(x,y)+\frac{1}{\gamma}\cdot\Delta(y,C_i)]
\end{eqnarray*}
For the right inequality,
\begin{eqnarray*}
d(x,y) &\le& \Delta(x,C_i)+\Delta(y,C_i)\\
&\le& \frac{1}{\gamma}\cdot\Delta(x,C_j)+\Delta(y,C_i)\\
&\le& \frac{1}{\gamma}\cdot(d(x,y)+\Delta(y,C_j))+\Delta(y,C_i)\\
&\le& \frac{1}{\gamma}\cdot(d(x,y)+\frac{1}{\gamma}\cdot\Delta(y,C_i))+\Delta(y,C_i)
\end{eqnarray*}
For part \ref{internal:part-2},
\begin{eqnarray*}
d(x,y) &\le& \Delta(x,C_i)+\Delta(y,C_i)\\
&\le& \frac{1}{\gamma}\cdot[\Delta(x,C_j)+\Delta(y,C_j)]\\
&\le& \frac{1}{\gamma}\cdot[2\cdot\Delta(x,C_j)+d(x,y)]
\end{eqnarray*}
$\square$

Note that for $\gamma \to \infty$ all distances $d(x,y)$ with $x \in C_i$ and $y \in C_j$ are roughly equal and $d(x_1, x_2) \ll d(x_1, y)$ for all $x_1, x_2 \in C_i$ and $y \in C_j$ with $i \neq j$.

We show next how to recover an $(\alpha,\gamma)$-clustering by sampling. For $x \in X$ and $A\subseteq X$ finite, we denote the average
distance from $x$ to $A$'s elements by $\Delta_U(x,A):= \frac{1}{|A|} \sum_{y \in A}d(x,y)$. A finite sample set $S$ provides us with an estimate for the distance of a point $x$ from a (not necessarily finite) $C \subseteq X$. Namely, we define the {\bf empirical proximity} of $x$ to $C$ as $\Delta_{emp}(x,C):=\Delta_U(x,C \cap S)$.

We turn to explain how we recover an unknown $(\alpha,\gamma)$-clustering of $X$ with $\alpha > 0$ and $\gamma > 1$. Consider a collection
$\cal{C}\rm=\{C_1,\ldots,C_k\} \subseteq X$ of disjoint subsets of $X$.
We define a ``Voronoi diagram" corresponding to $S$, denoted $\mathcal C^\gamma=\{C_1^\gamma,\ldots,C_k^\gamma\}$. Here

$$C_{i}^\gamma=\{x\in X:\forall j\ne i,\;\gamma\cdot\Delta_{emp}(x,C_i)<\Delta_{emp}(x,C_j)\}.$$
If $\cal C$ is a $(\alpha,\gamma)$-clustering of $X$, we expect $\mathcal C^\gamma$ to be a good approximation of $\cal{C}$.

\begin{lemma}\label{lemma:sampling-lemma}
Let $\mathcal C=\{C_1,\ldots,C_k\}$ be an $(\alpha,\gamma)$-clustering of $X$. Let $S=\{Z_1,\ldots,Z_m\}$ be an i.i.d. sample with distribution $P$ and let $q\ne p$. Then, for every $x\in C_q,\epsilon>0$,
$$P\left(\Delta_{emp}(x,C_p)\ge (\gamma-\epsilon)\cdot\Delta_{emp}(x,C_q)\right)\ge
1-3\exp\left(-\left(\frac{\epsilon(\gamma-1)\alpha}{\sqrt{8}\gamma(\gamma^2+1)}\right)^2\cdot m\right)$$
\end{lemma}
The proof follows by a standard application of the Hoeffding bound and is deferred to the appendix.
\begin{corollary}\label{cor:clustering-can-be-approximated}
Let $S=\{Z_1,\ldots,Z_m\}$ be an i.i.d. sample with distribution $P$. Then, for every $(\alpha,\gamma)$-clustering $\mathcal C$, $\Pr(d(\mathcal C,\mathcal C^{\gamma -\delta})>t)\le \frac{3}{t\alpha}\cdot\exp\left(-\left(
\frac{(\gamma-1)\delta\alpha}{\sqrt{8}\gamma(\gamma^2+1)}\right)^2\cdot m\right)$.	
\end{corollary}
{\bf Proof.} Denote $\mathcal C=\{C_1,\ldots,C_k\}$. By lemma \ref{lemma:sampling-lemma},  with $\epsilon=\delta$, we have
\begin{eqnarray*}
E[d(C,C^{\gamma -\delta})] &=& E[P(\cup_{i=1}^k C_i\oplus C^{\gamma -\delta}_{i})]\\
&=&\sum_{i=1}^k\int_{C_i}\Pr(x\notin C^{\gamma -\delta}_{i})dP(x)\\
&=&\sum_{i=1}^k\sum_{j\neq i}\int_{C_i}\Pr(x\in C^{\gamma -\delta}_{j})dP(x)\\
&\le&\sum_{i=1}^k(k-1)\cdot P(C_i)\cdot 3\cdot\exp\left(-\left(
\frac{(\gamma-1)\delta\alpha}{\sqrt{8}\gamma(\gamma^2+1)}\right)^2\cdot m\right)\\
&=& (k-1)\cdot 3\cdot\exp\left(-\left(
\frac{(\gamma-1)\delta\alpha}{\sqrt{8}\gamma(\gamma^2+1)}\right)^2\cdot m\right)
\end{eqnarray*}
Thus, the lemma follows from Markov's inequality and the fact that $k-1\le k\le \frac{1}{\alpha}$
$\square$

We next turn to investigate the collection of all $(\alpha,\gamma)$-clusterings of the given space. 
We observe first that every two distinct $(\alpha,\gamma)$-clusterings must differ substantially.

\begin{lemma}\label{lemma:distinct-clusterings-are-far}
If $\mathcal C,\mathcal C'$ are two $(\alpha,\gamma)$-clusterings with $d(\mathcal C,\mathcal C')>0$, then 
$d(\mathcal C,\mathcal C')\ge\frac{\alpha\cdot(\gamma-1)^2}{2\gamma^2-\gamma+1}.$
\end{lemma}
{\bf Proof.} Denote $\mathcal C=\{C_1,\ldots,C_k\},\;\mathcal C'=\{C'_1,\ldots,C'_{k'}\}$ and $\epsilon=d(\mathcal C,\mathcal C')$. By adding empty clusters if needed, we can assume that $k=k'$. By reordering the clusters,
if necessary, we can assume that $P(\cup_{i=1}^{k}C_i\oplus C'_i)=\epsilon$
and $P(C_1'\oplus C_1)>0$. Again by selecting the ordering we can assume the existence of some point $x$ that is in $C_1'\setminus C_1$ and in $C_2\setminus C_2'$.
\begin{eqnarray}\label{eq:1}
\Delta(x,C_1') &=& \frac{1}{P(C_1')}\cdot\int_{C_1'}d(x,y)dP(y)\nonumber\\
&\ge& \frac{1}{P(C_1')}\cdot\int_{C_1}d(x,y)dP(y)-\frac{1}{P(C_1')}\cdot\int_{C_1\setminus C_1'}d(x,y)dP(y)\nonumber\\
&\ge& \frac{P(C_1)}{P(C_1')}\cdot\Delta(x,C_1)-\frac{P(C_1\setminus C_1')}{\alpha}\cdot\max_{y\in C_1\setminus C_1'}d(x,y)\\
&\ge&\left(1-\frac{\epsilon}{\alpha}\right)\cdot\Delta(x,C_1)
-\frac{\epsilon}{\alpha}\cdot\frac{\gamma^2+1}
{\gamma(\gamma-1)}\Delta(x,C_1)\nonumber\\
&\ge& \left(1-\frac{\epsilon}{\alpha}\cdot\frac{2\gamma^2-\gamma+1}
{\gamma(\gamma-1)}\right)\cdot\gamma\cdot\Delta(x,C_2)\nonumber
\end{eqnarray}
For the second inequality note that $\frac{P(C_1')}{P(C_1)}\ge \frac{P(C_1)-P(C_1\setminus C_1')}{P(C_1)}\ge 1-\frac{\epsilon}{\alpha}$. The third inequality follows from lemma \ref{lemma:dist-bound}.

As we just saw $\frac{\Delta(x,C_1')}{\Delta(x,C_2)}\ge \left(1-\frac{\epsilon}
{\alpha}\cdot\frac{2\gamma^2-\gamma+1}
{\gamma(\gamma-1)}\right)\cdot\gamma$.
The same argument yields as well $\frac{\Delta(x,C_2)}{\Delta(x,C_1')}\ge\left(1-\frac{\epsilon}
{\alpha}\cdot\frac{2\gamma^2-\gamma+1}
{\gamma(\gamma-1)}\right)\cdot\gamma$. Consequently $1 \ge \left(1-\frac{\epsilon}
{\alpha}\cdot\frac{2\gamma^2-\gamma+1}
{\gamma(\gamma-1)}\right)\cdot\gamma$
which proves the lemma. $\square$

As we observe next, for every $\alpha > 0$ and $\gamma > 1$ the number of $(\alpha,\gamma)$-clusterings that
any space can have does not exceed $f(\alpha,\gamma)$, where $f$ depends only on $\alpha$ and $\gamma$ but {\it not} on the space. We find this somewhat surprising, although the proof is fairly easy. 

\begin{corollary}\label{only_finite_number}
There is a function $f=f(\alpha,\gamma)$ defined for $\alpha > 0$ and $\gamma > 1$ with the following property. The number of $(\alpha,\gamma)$-clusterings of {\em any} metric probability space $X$ is at most $f(\alpha,\gamma)$. This works in particular with
$f(\alpha,\gamma)=2\cdot\left(\frac{12(2\gamma^2-\gamma+1)}
{\alpha^2(\gamma-1)^2}\right)^{\left(\frac{\sqrt{8}\gamma(\gamma^2+1)}
{(\gamma-1)^2\alpha}\right)^2\cdot\ln(\frac{1}{\alpha})}$
\end{corollary}
{\bf Proof.}
Consider the following experiment.
We take an i.i.d. sample $Z_1,\ldots,Z_m$ of points from the distribution $P$ with

$$m>\left(\frac{\sqrt{8}\gamma(\gamma^2+1)}
{(\gamma-1)^2\alpha}\right)^2\cdot\ln\left(\frac{12(2\gamma^2-\gamma+1)}
{\alpha^2(\gamma-1)^2}\right).$$
and partition them randomly into $k\le (\frac{1}{\alpha})$ parts $T_1,\ldots,T_k$. This induces a partition $\mathcal C^*=\{C_1,\ldots,C_k\}$ of the space $X$ defined by
$$C_{i}=\{x\in X:\forall j\ne i,\;\Delta_{U}(x,T_i)<\Delta_{U}(x,T_j)\}$$
For every $(\alpha,\gamma)$-clustering $\mathcal C$ of $X$ we consider the event $A_{\mathcal C}$ that the induced partition of $X$ satisfies $d(\mathcal C,\mathcal C^*)<\alpha\cdot\frac{(\gamma-1)^2}{2\cdot(2\gamma^2-\gamma+1)}$.
Let us consider the events $A_{\mathcal C}$ over distinct $(\alpha,\gamma)$-clusterings of the space.
By Lemma \ref{lemma:distinct-clusterings-are-far}, these events $A_{\mathcal C}$ are disjoint.
Now consider the event $B$ that the $T_i$'s are consistent with $\cal C$. There are at most $(\frac{1}{\alpha})^m$ ways to partition the sampled points into $\frac{1}{\alpha}$ parts or less, so that $\Pr(B)\ge \alpha^m$.  By the choice of $m$ and by Corollary \ref{cor:clustering-can-be-approximated} $\Pr(A_{\mathcal C}|B)\ge \frac 12$. Thus, $\Pr(A_{\mathcal C})\ge\Pr(B)\cdot \Pr(A_{\mathcal C}|B)\ge \frac 12 \alpha^m$.
Consequently, $X$ has at most $f(\alpha,\gamma)=2(\frac{1}{\alpha})^m$ distinct $(\alpha,\gamma)$-clusterings, as claimed. $\square$

\begin{note}\label{note:many_clusters} 
Fix $\alpha>0$. The number of $(\alpha,\gamma)$-clusterings might be quite large when $\gamma$ is close to $1$. For example, let $X$ be an $n$-point space, with uniform metric and uniform probability measure. Every partition in which each part has cardinality $\ge \alpha \cdot n$ is an $(\alpha,\frac{n}{n-1})$-clustering\footnote{Note that this example is not valid if we define $\Delta(x,A)=E[d(x,y)|y\in A\setminus\{x\}]$. To overcome this point, we can replace every point $x \in X$ by many copies, where two copies of $x$ are distance $\epsilon$ and a copy of $x$ and a copy of $y \neq x$ are at distance $d(x,y)$.}.
\end{note}

\subsection*{Algorithmic Aspects}
Fix $\alpha>0,\gamma>1$. We shall now show that an $(\alpha,\gamma)$-clustering can be well approximated efficiently. By lemma \ref{cor:clustering-can-be-approximated}, $(\alpha,\gamma)$-clustering can be approximated by a small sample, where the approximation is with respect to the symmetric difference metric. A major flaw of this approximation scheme is that we have no verification method to accompany it. We do not know how to check whether a given partition is close to an $(\alpha,\gamma)$-clustering w.r.t. the symmetric difference metric. To this end, we introduce another notion of approximation. A family of subsets of $X$, $\mathcal C=\{C_1,\ldots,C_k\}$, is an {\bf $(\epsilon,\alpha,\gamma)$-clustering} if 
\begin{itemize}
\item For every $i\in [k]$, $P(C_i)\ge \alpha$
\item There is a set $N\subset X$ with $P(N)\le \epsilon$ such that every $x\in X\setminus N$, belongs to exactly one $C_i$ and for every $j\ne i$, $\Delta(x,C_j)\ge\gamma\cdot\Delta(x,C_i)$.
\end{itemize}

We consider next a partition that is attained by the method of Corollary~\ref{cor:clustering-can-be-approximated}. We show that if it is $\epsilon$-close to an $(\alpha,\gamma)$-clustering w.r.t. symmetric differences, then it is necessarily an $(\alpha-\epsilon,\gamma-O(\epsilon),\epsilon)$-clustering. 

We associate with every collection $\mathcal A =\{A_1,\ldots,A_k\}$ of finite subsets\footnote{In fact, we will allow $A_1,\ldots,A_k$ to have multiple points. Formally, then, $A_1,\ldots,A_k$ are multisets.} of $X$ the following collection of subsets $\mathcal C^\gamma(\mathcal A)=\{C_1^\gamma(\mathcal A),\ldots,C_k^\gamma(\mathcal A)\}$:
\begin{equation}
\label{induced:partition}
C_{i}^\gamma(\mathcal A)=\{x\in X:\forall j\ne i,\;\gamma\cdot\Delta_{U}(x,A_i)<\Delta_{U}(x,A_j)\}
\end{equation}
where, as above, $\Delta_U(x,A):=\frac{1}{|A|}\sum_{z\in A}d(x,z)$.

\begin{proposition}\label{prop_sym_dist_entails_approx}
Let $\mathcal C=\{C_1,\ldots,C_k\}$ be an $(\alpha,\gamma)$-clustering. Let $\mathcal A=\{A_1,\ldots, A_k\}$ where $\forall i,\;A_i\subset C_i$ and $d(\mathcal C^\gamma(\mathcal{A}),\mathcal C)<\epsilon$. Then $\mathcal C^\gamma(\mathcal{A})$ is an $(\alpha-\epsilon ,\gamma-O(\epsilon),\epsilon)$-clustering. The unspecified coefficients in the $O$-term depend on $\alpha$ and $\gamma$.
\end{proposition}
The main idea of the proof is rather simple: The assumption $d(\mathcal C^\gamma(\mathcal A),\mathcal C)<\epsilon$ implies that for all $i$ the set $C_i\oplus C_i^\gamma(\mathcal A)$ is small. This suggests that $\Delta(x,C_i)\approx \Delta(x,C_i^\gamma(\mathcal A))$ for most points $x\in  X$. The only difficulty in realizing this idea is that points in $C_i\oplus C_i^\gamma(\mathcal A)$ might have a large effect on either $\Delta(x,C_i)$ or $\Delta(x,C_i^\gamma(\mathcal A))$. But the assumption that $A_i\subset C_i$ gives us control over the distances between $x$ to these points. The full proof can be found in the appendix.

To recap, the above discussion suggests a randomized algorithm that for a given $\epsilon > 0$ runs in time $\mbox{poly}(\frac{1}{\epsilon})$ and finds w.h.p. an $(\alpha-\epsilon,\gamma-O(\epsilon),\epsilon)$-clustering of $X$ provided that $X$ has an $(\alpha,\gamma)$-clustering $\mathcal{C}$. We take $m = \Theta(\log(\frac{1}{\epsilon}))$ i.i.d. samples from $X$ and go over all possible partitions of the sample points into at most $\frac{1}{\alpha}$ sets. There are only $\left(\frac{1}{\epsilon}\right)^{O(\log(\frac{1}{\alpha}))}$ such partitions.
We next check whether the clustering of $X$ that is induced as
in Equation~(\ref{induced:partition})
is an $(\alpha-\epsilon,\gamma-O(\epsilon),\epsilon)$-clustering (this can be easily done by standard statistical estimates). 

To see that the algorithm accomplishes what it should, note that the failure probability in corollary \ref{cor:clustering-can-be-approximated} with $\delta=\gamma-1$ can be $\le \frac12$ for $m = \Theta(\log(\frac{1}{\epsilon}))$. Thus, w.p. $>\frac12$ one of the considered partitions induces a partition of $X$ which is $\epsilon$-close in the symmetric difference sense to $\mathcal{C}$. By Proposition \ref{prop_sym_dist_entails_approx}, this partition is an $(\alpha-\epsilon,\gamma-O(\epsilon),\epsilon)$-clustering.

This also proves Theorem \ref{taste}: If our input is a finite metric space $X$, we can apply the above algorithm with $\epsilon=\frac{1}{|X|+1}$ and examples that are being sampled from $X$ uniformly at random. As explained, w.h.p., the algorithm will consider every partition which is $\epsilon$-close in the symmetric difference sense to any of $X$'s $(\alpha,\gamma)$-clusterings. However, since $\epsilon=\frac{1}{|X|+1}$, two $\epsilon$-close partitions must be identical. This proves Theorem \ref{taste}.

Note that by corollary \ref{only_finite_number}, {\em all} the $(\alpha,\gamma)$-clusterings can be approximated. A similar algorithm can efficiently find an approximate $(\alpha,\gamma,\epsilon)$-clustering, provided that one exists\footnote{The main difference is that here we do not consider partitions of the whole sample set. Rather, we seek first those sample points that belong to the exceptional set, and only partitions of the remaining sample points are considered.}. Also, similar techniques yield an algorithm to approximate an individual $(\alpha,\gamma)$-cluster.

\section{Clustering into Many Clusters}\label{sec:many_clusters}
To simplify matters we consider only finite metric spaces endowed with a uniform probability distribution\footnote{As in the previous section, it's a fairly easy matter to accommodate general metric spaces and arbitrary probability distributions.}.
\begin{lemma}\label{lemma:3-clusters}
Let $X$ be a metric space and let $\epsilon>0$.
\begin{enumerate}
\item Let $C_1,C_2\subseteq X$ be two $(3+\epsilon)$-clusters. Then $C_1\cap C_2=\emptyset,\; C_1\subset C_2$ or $C_2\subset C_1$.
\item Every $(3+\epsilon)$-cluster is a ball around one of its points.
\item
The claim is sharp and the above claims need not hold for $\epsilon=0$.
\end{enumerate}
\end{lemma}
{\bf Proof.} We prove the first claim by contradiction and assume that $P(C_1\setminus C_2),P(C_2\setminus C_1),P(C_1\cap C_2)$ are positive. Let $x\in C_1\cap C_2$, $y\in C_1\oplus C_2$ be such that $d(x,y)$ is as small as possible. Say that $y\in C_2$. Clearly, $\Delta(x,C_1\setminus C_2)\ge d(x,y)$.

We first deal with the case $P(C_1\setminus C_2)\ge P(C_1\cap C_2)$, and arrive at a contradiction as follows:
\begin{eqnarray*}
\Delta(x,C_1) &=& \frac{P(C_1\setminus C_2)}{P(C_1)}\Delta(x,C_1\setminus C_2)+\frac{P(C_1\cap C_2)}{P(C_1)}\Delta(x,C_1\cap C_2)\\
&\ge& \frac{1}{2}\Delta(x,C_1\setminus C_2)\\
&\ge& \frac{1}{2}d(x,y)\\
&\ge& \frac{1}{2}[\Delta(y,C_1)-\Delta(x,C_1)]\\
&\ge& \frac{3+\epsilon-1}{2}\Delta(x,C_1)
\end{eqnarray*}
When $P(C_1\setminus C_2)\le P(C_1\cap C_2)$, a contradiction is reached as follows. By the choice of $x,y$, for every $z\in C_1\setminus C_2$, there holds $\Delta(z,C_1\cap C_2) \ge d(x,y)$. Therefore,
\begin{eqnarray*}
\Delta(z,C_1) &=& \frac{P(C_1\setminus C_2)}{P(C_1)}\Delta(z,C_1\setminus C_2)+\frac{P(C_1\cap C_2)}{P(C_1)}\Delta(z,C_1\cap C_2)\\ 
&\ge& \frac{1}{2}\Delta(z,C_1\cap C_2)\\
&\ge& \frac{1}{2}d(x,y)\\
&\ge& \frac{1}{2}[\Delta(y,C_1)-\Delta(x,C_1)]\\
&\ge& \frac{1}{2}\cdot(1-\frac{1}{3+\epsilon})\cdot \Delta(y,C_1)\\
&\ge&\frac{1}{2}\cdot(1-\frac{1}{3+\epsilon})\cdot(3+\epsilon)\cdot \Delta(z,C_1)
\end{eqnarray*}
To prove the second part, let $C$ be a $(3+\epsilon)$-cluster of diameter $r$, and let $x,y\in C$ satisfy $d(x,y)=r$. Since $d(x,y)\le \Delta(x,C)+\Delta(y,C)$, we may assume w.l.o.g. that $\Delta(x,C)\ge \frac{d(x,y)}{2}$. We show now that $C=B(x,r)$ and $C$ is a ball, as claimed. Indeed $d(x,z)\le r$ for every $z\in C$, and if $z\notin C$, then $d(x,z)\ge \Delta(z,C)-\Delta(x,C)\ge (3+\epsilon-1)\Delta(x,C)>d(x,y)=r$. The conclusion follows.

To show that the result is sharp, consider the graph $G$ that is a four-vertex cycle and its graph metric. It is not hard to check that every two consecutive vertices in $G$ constitute a $3$-cluster which is not a ball. Moreover a pair of intersecting edges in $G$ yield an example for which the first part of the lemma fails to hold.
$\square$

An $(\alpha, \gamma)$-cluster in a space $X$ is called {\em minimal} if it contains no $(\alpha, \gamma)$-cluster other than itself. Such clusters are of interest, since they can be viewed as ``atoms'' in clustering $X$.
\begin{corollary}
For every $\alpha, \epsilon > 0$ and every space $X$
there is at most one partition of $X$ into minimal $(\alpha,3+\epsilon)$-clusters.
\end{corollary}
To see this, consider two $(\alpha,3+\epsilon)$-clusters $C$ and $C'$ that belong to two different such partitions and have a nonempty intersection. By Lemma \ref{lemma:3-clusters}, they must be comparable. By the minimality assumption, $C=C'$ which proves the claim.

\begin{note}
We note that the previous Corollary may fail badly without the minimality assumption. Let $X=\{x_1,\ldots,x_n\} \dot \cup \{y_1,\ldots,y_n\}$, where $d(x_i, y_i)=1$ for all $i$ and all other distance equal $\gamma$. It is not hard to
see that the following are $(\alpha, \gamma)$-clusters in $X$ where $\alpha = \frac{1}{2n}$: A singleton and a pair $\{x_i,y_i\}$. There are 
$2^n=2^{\frac{1}{2\alpha}}$ ways to partition $X$ into such clusters.  
\end{note}

\subsection*{Algorithmic Aspects}
We next discuss several algorithmic aspects of clustering into arbitrarily many clusters. Our input consists of a finite metric space $X$ and the parameter $\alpha > 0$. Lemma \ref{lemma:3-clusters} suggests an algorithm for finding $(\alpha,3+\epsilon)$-clusters and for partitioning the space into $(\alpha,3+\epsilon)$-clusters. The runtime of this algorithm is polynomial in $|X|$, and independent of $\alpha$. The second part of the lemma suggests how to find all the $(\alpha,3+\epsilon)$-clusters. As the first part of the lemma shows, the inclusion relation among the  $(\alpha,3+\epsilon)$-clusters has a tree structure. Thus, we can use dynamic programming to find a partition of the space into  $(\alpha,3+\epsilon)$-clusters, provided that such a partition exists. This proves the positive part of Theorem \ref{dessert}.

To match the above positive result, we show
\begin{theorem}\label{th:NP-Hardness}
The following problems are NP-Hard. 
\begin{enumerate}
\item $(\alpha,2.5)$-CLUSTERING: Given an $n$-point metric space $X$ and $\alpha>0$, decide whether $X$ has a $(\alpha,2.5)$-clustering.
\item PARTITION-INTO-$(\alpha,2.5)$-CLUSTERS: Given an $n$-point metric space $X$ and $\alpha>0$, decide whether $X$ has a partition into $(\alpha,2.5)$-clusters.
\end{enumerate}
\end{theorem}
The proof of this Theorem, which also proves the negative part of Theorem \ref{dessert}, is deferred to the appendix.

\section{Conclusion}\label{sec:conclusion}
\subsection{Relation to other work}
As we explain below, our work is  inspired
by the classical VC/PAC theory. In addition we refer to several recent papers that contribute to the development of a theory of clustering.

\subsubsection*{VC/PAC theory}
The VC/PAC setting offers the following formal description of the classification problem. We are dealing with a space $\mathcal X$ of {\em instances}. The problem is to recover an unknown member $h^{*}$ in a known class $\mathcal H$ of hypotheses. Here 
$\mathcal H\subset\mathcal{Y}^\mathcal{X}$, where $\cal Y$ is a finite set of {\em labels}. We seek to recover the unknown $h^{*}$ by observing a sample $S=\{(x_i,h^{*}(x_i)\}_{i=1}^m\subset\mathcal{X}\times\mathcal{Y}$. These samples come from some fixed but unknown distribution over $\mathcal X$.

Our description of the clustering problem is similar. We consider a space $X$ of instances and a class $\mathcal G$ of good clusterings $(P,\mathcal C)$ of $X$, where $P$ is probability measure over $X$ and $\mathcal C$ is a partition of $X$. We are given a sample $\{X_1,\ldots,X_m\}\subset X$ that comes from some unknown $P$, where $(P,\mathcal C)\in \mathcal G$ for some partition $\mathcal C$, and our purpose is to recover $\mathcal C$.
Specifically, here $X$ is a metric space, $\mathcal G$ is the class of probability measures $P$ that admit a partition which is a $(\alpha,\gamma)$-clustering and the corresponding partition is the associated $(\alpha,\gamma)$-clustering. 

Both theories seek conditions on $\mathcal G$ or $\mathcal H$ under which there are no information theoretic or computational obstacles that keep us from performing the above mentioned tasks.

\subsubsection*{Alternative Notions of Good Clustering}

Our approach is somewhat close in spirit to \cite{BalcanBlVem08}, see also
\cite{Blum09}. These papers assume that the space under consideration has a clustering with some structural properties, and show how to find it efficiently. In particular, a key notion in these papers is the {\bf $\gamma$-average attraction property}, which is conceptually similar to our notion of $\gamma$-clustering. Given a partition $\mathcal C=\{C_1,\ldots,C_k\}$ of a space $X$ it is possible to compare between clusters either additively or through multiplication. In \cite{BalcanBlVem08} the requirement is that $\Delta(x,C_i)+\gamma\le \Delta(x,C_j)$ for every $x\in C_i$ and $j\ne i$, whereas our condition is $\Delta(x,C_i)\cdot\gamma\le \Delta(x,C_j)$. A clear advantage of our notion is its scale invariance. On the other hand, their algorithms work even if $X$ is not a metric space and is only endowed with an arbitrary dissimilarity function.

We mention two more papers that share a similar spirit. Consider a data set that resides in the unit ball of a Hilbert Space. It is shown in \cite{KarLibLovSchWein11} how to efficiently find a large margin classifier for the data provided that one exists. In \cite{AckBen-David08} several additional possible notions of good clustering are introduced and analyzed.

\subsubsection*{Stability}
The notion of instance stability was introduced in \cite{BilLin10} (See also \cite{AwasthiBlumSh11}). An instance for an optimization problem in called {\em stable} if the optimal solution does not change (or changes only slightly) upon a small perturbation of the input. The point is made that instances of clustering problems are of practical interest only if they are stable. The notion of an $(\alpha,\gamma)$-clustering has a similar stability property. Namely, if we slightly perturb a metric, an $(\alpha,\gamma)$-clustering is still $(\alpha',\gamma')$-clustering for $\alpha'\approx,\alpha,\;\gamma'\approx \gamma$.  Thus, a good clustering remains a good clustering under a slight perturbation of the input

In fact, the present paper is an outgrowth of our work on stable instances for MAXCUT, which we view as a clustering problem. We recall that the input to the MAXCUT problem is an $n \times n$ nonnegative symmetric matrix $W$. We seek an $S \subseteq [n]$ which maximizes $\sum_{i \in S, j \not\in S} w_{ij}$. Even METRIC-MAXCUT problem (i.e., when $w_{ij}$ form a metric) is $NP$-Hard~\cite{}. We say that $W'$ is a $\gamma$-perturbation of $W$ some $\gamma>1$ if $\forall i,j,\;\gamma^{-\frac{1}{2}} w_{ij}\le w'_{i,j}\le \gamma^{\frac{1}{2}}w_{ij}$.
The instance $W$ of MAXCUT is called {\bf $\gamma$-stable} if the  optimal solution $S$ for $W$ coincides with the optimal solution for every $\gamma$-perturbation $W'$ of $W$. The methods presented in this paper can be used to give, for every $\epsilon>0$, an efficient algorithm that correctly solves all
$(1+\epsilon)$-stable instances of METRIC-MAXCUT.

These developments will be elaborated in a future publication.

\subsection{Future Work and Open Questions}

In view of this article and papers such as   \cite{AckBen-David08,KarLibLovSchWein11,BalcanBlVem08} it is clear that there is still much interest in new notions of a good clustering and the relevant algorithms. Still, on the subjects discussed here several natural questions remain open.
\begin{enumerate}
\item
We believe that it should be possible to improve the dependence on $\alpha$ and $\gamma$ of the run time of the algortihm in Theorem~\ref{taste}.
\item We gave an efficient method for partitioning a space into $3$-clusters, and showed (theorem \ref{th:NP-Hardness}) that it is $NP$-Hard to find a partition into $2.5$-clusters. Can this gap be closed?
\item
As Lemma~\ref{lemma:3-clusters} shows, $(3+\epsilon)$-clusters are just balls. It is not hard to see that Lemma \ref{lemma:sampling-lemma} implies that given an $(\alpha,\gamma)$-clustering of an $n$-point metric space, it is possible to find $O_{\gamma}(\log n)$ {\em representative} points in every cluster so that the clustering is nothing but the Voronoi diagram of the (bunched) representative sets. Presumably, there is still some interesting structural theory of $(\alpha,\gamma)$-clustering waiting to be discovered here. Specifically, can the above $O_{\gamma}(\log n)$ be replaced by $O_{\gamma}(1)$? A positive answer would give a deterministic version of our algorithm from section \ref{sec:few-clusters}, with no dependency of $\alpha$, but only on the maximal number of clusters.
\item Consider the following statement ``Every $n$-point metric space $X$ has a partition $X=A\dot\cup B$ such that for every $x\in A,y\in B$, it holds that $\gamma(n)\cdot\Delta(x,A)\le \Delta(x,B)$ and $\gamma(n)\cdot\Delta(y,B)\le \Delta(y,A)$". How large can $\gamma(n)$ be for this statement to be true?
\end{enumerate}

\bibliographystyle{plain}
\bibliography{bib}
\appendix
\section{Proofs omitted from the text}
{\bf Proof.} (of Lemma \ref{lemma:sampling-lemma})
For $A\subset X$, denote $I_A=\frac 1m|\{j:Z_j\in A\}|,\;J_A=\sum_{j:Z_j\in A}d(x,Z_j)$. For every $j\in[m]$ define
\begin{equation*}
Y_j=
\begin{cases}
\frac{1}{P(C_p)}\cdot d(x,Z_j) & Z_j\in C_p\\
-\frac{\gamma-\frac{\epsilon}{2}}{P(C_q)}\cdot d(x,Z_j) & Z_j\in C_q\\
0 & otherwise
\end{cases}
\end{equation*}
We have $EY_j=\Delta(x,C_p)-(\gamma-\frac{\epsilon}{2})\cdot\Delta(x,C_q)\ge \frac{\epsilon}{2\gamma}\cdot\Delta(x,C_p)$. Moreover, by lemma \ref{lemma:dist-bound}, $|Y_j|\le 
\frac{(\gamma-\frac{\epsilon}{2})}{\alpha}\cdot
\frac{\gamma^2+1}{\gamma(\gamma-1)}
\cdot\Delta(x,C_p)
\le \frac{\gamma^2+1}{\alpha(\gamma-1)}
\cdot\Delta(x,C_p)$. Thus, by Hoeffding's bound,
$$P\left(\frac{J_{C_p}}{P(C_p)}\le \left(\gamma-\frac{\epsilon}{2}\right)\cdot
\frac{J_{C_q}}{P(C_q)}\right)=
P\left(\sum_{j=1}^mY_j\le 0 \right)\le \exp\left(-\left(\frac{\epsilon(\gamma-1)\alpha}{\sqrt{8}\gamma(\gamma^2+1)}\right)^2\cdot m\right)$$
Again by Hoeffding's bound, we have
$$P\left(\frac{I_{C_q}}{P(C_q)}\le 1-\frac{\epsilon}{4\gamma}\right)\le \exp\left(-\left(\frac{\epsilon\alpha}{\sqrt{8}\gamma}\right)^2\cdot m\right)$$
$$P\left(\frac{I_{C_p}}{P(C_p)}\ge 1+\frac{\epsilon}{4\gamma}\right)\le \exp\left(-\left(\frac{\epsilon\alpha}{\sqrt{8}\gamma}\right)^2\cdot m\right)$$
Combining the inequalities, we conclude that, with probability $\ge$ $1-3\exp\left(-\left(\frac{\epsilon(\gamma-1)\alpha}
{\sqrt{8}\gamma(\gamma^2+1)}\right)^2\cdot m\right)$,
\begin{eqnarray*}
\frac{\frac{J_{C_p}}{I_{C_p}}}{(\gamma-\epsilon)\frac{J_{C_q}}{I_{C_q}}}&=&
\frac{\frac{J_{C_p}}{P(C_p)}}{(\gamma-\frac{\epsilon}{2})\frac{J_{C_q}}{P(C_q)}}\cdot
\frac{\frac{P(C_p)}{I_{C_p}}}{\frac{P(C_q)}{I_{C_q}}}\cdot
\frac{\gamma-\frac{\epsilon}{2}}{\gamma-\epsilon}\\
&\ge&\frac{1-\frac{\epsilon}{4\gamma}}{1+\frac{\epsilon}{4\gamma}}\cdot
\frac{\gamma-\frac{\epsilon}{2}}{\gamma-\epsilon}\ge 1
\end{eqnarray*}
$\square$

\begin{proof} (of Proposition \ref{prop_sym_dist_entails_approx})
It is very suggestive how to select the exceptional set in the $(\alpha-\epsilon ,\gamma-O(\epsilon),\epsilon)$-clustering that we seek. Namely,
let $N=\cup_i \left(C_i\setminus C_{i}^\gamma(\mathcal{A})\right)$. As needed, $P(N)<\epsilon$, since $d(\mathcal C^\gamma(\mathcal A),\mathcal C)<\epsilon$. To prove our claim, note that $\forall i,\;P(C_{i}^\gamma)\ge \alpha-\epsilon$ since $d(\mathcal C,\mathcal C_*^\gamma(\mathcal{A}))<\epsilon$. Consider some $x\in X\setminus N$ and the unique index $i$ for which $x\in C_{i}^\gamma(\mathcal{A})\cap C_i$. If $j\ne i$, we need to show that
$$\Delta(x,C_{j}^\gamma(\mathcal A))\ge 
(\gamma - O(\epsilon))\Delta(x,C_{i}^\gamma(\mathcal A))$$
As in the proof of lemma \ref{lemma:distinct-clusterings-are-far}, we have
\begin{eqnarray}\label{eq:6}
\Delta(x,C_{j}^\gamma(\mathcal A)) &\ge & \left(1-\frac{\epsilon}{\alpha}\right)\Delta(x,C_j)-\frac{\epsilon}{\alpha}\max_{y\in C_j\setminus C_{j}^\gamma(\mathcal A)}d(x,y)\nonumber\\
&\ge & \left(1-\frac{\epsilon}{\alpha}\cdot\frac{2\gamma^2-\gamma+1}
{\gamma(\gamma-1)}\right)\cdot\Delta(x,C_j)\\
&=:&(1-a_1\cdot\epsilon)\cdot\Delta(x,C_j)\nonumber
\end{eqnarray}
Similarly, again as in the proof of lemma \ref{lemma:distinct-clusterings-are-far}, we have
\begin{equation}\label{eq:7}
\Delta(x,C_i)\ge \left(1-\frac{\epsilon}{\alpha}\right)\Delta(x,C_{i}^\gamma(\mathcal A))-\frac{\epsilon}{\alpha}\max_{y\in C_{i}^\gamma(\mathcal A)\setminus C_i}d(x,y)
\end{equation}
Now, for $y\in C_{i}^\gamma(\mathcal A)$, we have
\begin{eqnarray*}
d(x,y) &\le& \Delta_U(x,A_i)+\Delta_U(y,A_i)\\
&\le& \frac{1}{\gamma}\Delta_U(x,A_j)+\frac{1}{\gamma}\Delta_U(y,A_j)\\
&\le& \frac{2}{\gamma }\Delta_U(x,A_j)+ \frac{1}{\gamma}d(x,y)
\end{eqnarray*}
Now, since $A_j\subset C_j$, by lemma \ref{lemma:dist-bound}, $\Delta_U(x,A_j)\le  \frac{\gamma^2+1}{\gamma(\gamma-1)}\Delta(x,C_j)$ and we have,
$$d(x,y)\le \frac{\gamma}{\gamma-1}\cdot \frac{\gamma^2+1}{\gamma(\gamma-1)}\cdot\frac{2}{\gamma}\Delta(x,C_j)$$
So, by equation (\ref{eq:7}) we have,
\begin{equation}\label{eq:8}
\Delta(x,C_i)\ge (1-a_2\cdot \epsilon)\cdot\Delta(x,C_{i}^\gamma(\mathcal A))-a_3\cdot\epsilon
\cdot \Delta(x,C_j)
\end{equation}
For some positive constants $a_2,a_3$ which depend only of $\gamma$ and $\alpha$. Now by equations (\ref{eq:6}) and (\ref{eq:8}) we conclude that
\begin{eqnarray*}
\Delta(x,C_{j}^\gamma(\mathcal A))&\ge& (1-(a_1+\gamma a_3)\cdot\epsilon)\cdot\Delta(x,C_j)+
\gamma a_3\cdot\epsilon\cdot\Delta(x,C_j)\\
&\ge&(1-(a_1+\gamma a_3)\cdot\epsilon)\cdot\gamma\cdot\Delta(x,C_i)+\gamma a_3\cdot\epsilon\cdot\Delta(x,C_j)\\
&\ge& (1-(a_1+\gamma a_3)\cdot\epsilon)(1-a_2\cdot\epsilon)\gamma\cdot \Delta(x,C_{i}^\gamma(\mathcal A))\\
&=&(\gamma-O(\epsilon))\cdot \Delta(x,C_{i}^\gamma(\mathcal A))
\end{eqnarray*}

\end{proof}

{\bf Proof.} (of Theorem \ref{th:NP-Hardness}) Both claims are proved by the same reduction from 3-DIMENSIONAL-MATCHING (e.g., \cite{GareyJo79} pp. 221). The input to this problem is a subset $M\subset Y\times Z\times W$, where $Y,Z,W$ are three disjoint $q$-element sets.
A {\em three dimensional matching} (=3DM) is a $q$-element subset $M'\subset M$ 
that covers all elements in $Y\dot\cup Z\dot\cup W$. The problem is to decide whether a 3DM exists.  

We associate with this instance of the problem a graph on vertex set $Y\dot\cup Z\dot\cup W$, and edge set the union of all triangles $\{y,z,w\}$ over  $(y,z,w)\in M$. It is not hard to see that $3DM$ remains $NP$-Hard under the restriction that this graph is connected.

Here is our reduction. Given an instance $M\subset Y\times Z\times W$ of $3DM$, we construct a graph $G^M=(V^M,E^M)$ as follows: Associated with every $m=(y,z,w)\in M$ is a gadget below. We consider the clustering problem on $G^M$ with its natural graph metric.

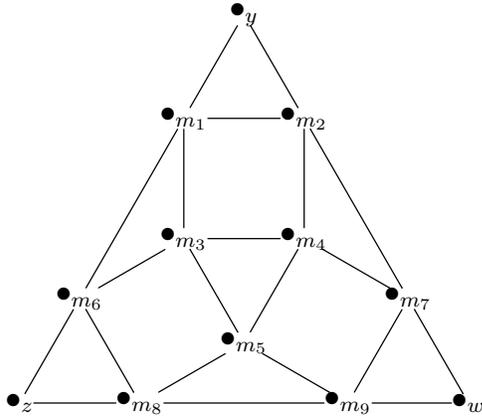
\xygraph{
!{<0cm,0cm>;<.8cm,0cm>:<0cm,.8cm>::}
!{(0,0) }*{\bullet_{y}}="y"
!{(-1,-1.73) }*{\bullet_{m_1}}="m_1"
!{(1,-1.73) }*{\bullet_{m_2}}="m_2"
!{(-1,-3.73) }*{\bullet_{m_3}}="m_3"
!{(1,-3.73) }*{\bullet_{m_4}}="m_4"
!{(0,-5.464) }*{\bullet_{m_5}}="m_5"
!{(-2.73,-4.73) }*{\bullet_{m_6}}="m_6"
!{(2.73,-4.73) }*{\bullet_{m_7}}="m_7"
!{(-1.73,-6.464) }*{\bullet_{m_8}}="m_8"
!{(1.73,-6.464) }*{\bullet_{m_9}}="m_9"
!{(-3.73,-6.464) }*{\bullet_{z}}="z"
!{(3.73,-6.464) }*{\bullet_{w}}="w"
"y"-"m_1","y"-"m_2","m_1"-"m_2","m_1"-"m_3","m_1"-"m_6",
"m_2"-"m_4","m_2"-"m_7","m_5"-"m_3","m_5"-"m_4","m_5"-"m_8","m_5"-"m_9",
"m_6"-"m_3","m_6"-"m_8","m_6"-"z","m_4"-"m_3",
"m_7"-"m_4","m_7"-"m_9","m_7"-"w","m_8"-"m_9","z"-"m_8","m_9"-"w"
}

We say that a triangle $T$ in a graph is {\em isolated} if every vertex outside it has at most one neighbor in $T$.
The above gadget is useful for the reduction since it's easy to verify that:
\begin{claim}\label{claim:2}
The graph $G^M$ can be partitioned into isolated triangles iff $M$ has a $3DM$.
\end{claim}

{\bf Proof(sketch).} If $M$ has a $3DM$, we can construct a partition of $V$ into isolated triangle by taking the triangles
\begin{equation}\label{eq:3}
\{y,m_1,m_2\},\{z,m_6,m_8\},\{w,m_7,m_9\},\{m_3,m_4,m_5\}
\end{equation}
for every $m$ in the $3DM$ and the triangles
\begin{equation}\label{eq:4}
\{m_1,m_3,m_6\},\{m_2,m_4,m_7\},\{m_5,m_8,m_9\}
\end{equation}
for $m$ outside it. On the other hand, consider any partition of $G^M$ into isolated triangles. Its restriction to every gadget must coincide with one of the above two choices, so that the corresponding $3DM$ is readily apparent
$\square$

Both $NP$-Hardness claims in Theorem~\ref{th:NP-Hardness} follow from the above discussion and the following claim
\begin{claim}\label{claim:1}
Let $G=(V,E)$ be a connected graph in which all vertex degrees are $\ge 2$. For every partition of the vertex set $V = \dot \cup_1^k C_i$, the following are equivalent
\begin{enumerate}
\item Each $C_i$ induces an isolated triangle.
\item Each $C_i$ is a $(\frac{3}{|V|},2.5)$-cluster.
\item The partition $C_1,\ldots,C_k$ is a $(\frac{3}{|V|},2.5)$-clustering.
\end{enumerate}
\end{claim}

{\bf Proof} The implication $1.\Rightarrow 2.$ and $\;1.\Rightarrow 3.$ are easily verified. We turn to prove $3.\Rightarrow 1$. Let $i\in[k]$. We need to show that each $C_i$ is an isolated triangle. Clearly, $|C_i|\ge 3$ by definition of $(\frac{3}{|V|},2.5)$-clustering. But $G$ is connected, so there are two neighbors $xy$ with $x\in C_i, y \not\in C_i$. By proposition \ref{lemma:dist-bound} we have
$$1=d(x,y)\ge (2.5-1)\Delta (x,C_i)\ge 1.5\cdot\frac{|C_i|-1}{|C_i|},$$
so that $|C_i|=3$. Consider now $x,y\in C_i$ which are nonadjacent. Since $d(x)\ge 2$, it has a neighbor $z\not\in C_i$. Using Proposition \ref{lemma:dist-bound} we arrive at the following contradiction:
$1=d(x,z)\ge (2.5-1)\Delta (x,C_i)\ge 1.5\cdot \frac {2+1}{3}=1.5$. We already know that each $C_i$ is a triangle, but why is it isolated? If $z \in C_j,\;j\ne i$ has at least two neighbors in $C_i$, then
$$2.5\cdot\Delta(z,C_j)\le \Delta(z,C_i)\le \frac{4}{3}<2.5\cdot \frac{2}{3}=2.5\cdot\Delta(z,C_j).$$

The proof of $2.\Rightarrow 1.$ is similar. Let $C_i$ be cluster in the partition. Using the same argument as before, where the the fact that $\forall x\in C_i,y\notin C_i,\;d(x,y)\ge \Delta(y,C_i)-\Delta(y,C_i)\ge (2.5-1)\Delta (x,C_i)$ replace Proposition \ref{lemma:dist-bound}, we deduce that $C_i$ induces a triangle. To show that $C_i$ is isolated, suppose that there exists a vertex $z\notin C_i$ with $\ge 2$ neighbors in $C_i$. Let $x\in C_i$ be an arbitrary vertex. To obtain a contradiction, we note that
$$2.5\Delta(x,C_i)\le \Delta(z,C_i)\le \frac{4}{3}<2.5\cdot \frac{2}{3}=2.5\Delta(x,C_i)$$
$\square$

\begin{note}
Theorem~\ref{th:NP-Hardness} is tight in the following sense: As the proof shows, the above problems are hard even for graph metrics. On the other hand, given a graph $G=(V,E)$, the following polynomial time algorithms find (i) A partition into $(\alpha,2.5+\epsilon)$-clusters, and (ii) A $(\alpha,2.5+\epsilon)$-clustering. (provided, of course, that one exists).
\begin{enumerate}
\item If $\alpha>\frac{1}{|V|}$ then, as in the proof of theorem \ref{th:NP-Hardness}, one shows that a partition into $(\alpha,2.5+\epsilon)$-clusters / $(\alpha,2.5+\epsilon)$-clustering is equivalent to a perfect matching, no edge of which is contained in a triangle. This can be done by first eliminating every edge that belongs to a triangle and then running an arbitrary matching algorithm.
\item If $\alpha < \frac{1}{|V|}$ then clearly there is no partition into $(\alpha,2.5+\epsilon)$-clusters / $(\alpha,2.5+\epsilon)$-clustering. If $\alpha = \frac{1}{|V|}$, the singletons constitute a partition of $V$ into $(\alpha,2.5+\epsilon)$-clusters and a $(\alpha,2.5+\epsilon)$-clustering.
\end{enumerate}
\end{note}

\begin{note}
As in Note \ref{note:many_clusters}, by replacing each vertex with many points at distance $\epsilon$ from each other, the above reduction applies as well with the definition $\Delta(x,A)=E[d(x,y)|y\in A\setminus\{x\}]$.
\end{note}
\end{document}